\documentclass[sigconf]{acmart}

\usepackage[utf8]{inputenc} 
\usepackage[T1]{fontenc}    
\usepackage{hyperref}       
\usepackage{url}           
\usepackage{booktabs}       
\usepackage{amsfonts}       
\usepackage{nicefrac}       
\usepackage{microtype}      
\usepackage{xcolor}         
\usepackage{graphicx}
\usepackage{algorithm}
\usepackage{algorithmic}
\usepackage{amsmath}
\usepackage{multirow}
\usepackage{colortbl}
\usepackage{color}
\usepackage{tabularray}
\usepackage{wrapfig}
\usepackage{subfloat}
\usepackage{graphicx}
\usepackage{subcaption}

\copyrightyear{2025}
\acmYear{2025}
\setcopyright{acmlicensed}\acmConference[KDD '25]{Proceedings of the 31st ACM SIGKDD Conference on Knowledge Discovery and Data Mining V.1}{August 3--7, 2025}{Toronto, ON, Canada}
\acmBooktitle{Proceedings of the 31st ACM SIGKDD Conference on Knowledge Discovery and Data Mining V.1 (KDD '25), August 3--7, 2025, Toronto, ON, Canada}
\acmDOI{10.1145/3690624.3709287}
\acmISBN{979-8-4007-1245-6/25/08}

\begin{document}
\title{SEPTQ: A Simple and Effective Post-Training Quantization Paradigm for Large Language Models}

\author{Han Liu}
\affiliation{
\institution{Dalian University of Technology}
\city{Dalian}
\country{China}}
\email{liu.han.dut@gmail.com}

\author{Haotian Gao}
\affiliation{
  \institution{Dalian University of Technology}
  \city{Dalian}
  \country{China}}
\email{haotian.dlut@gmail.com}

\author{Xiaotong Zhang}
\authornote{Corresponding author.}
\affiliation{
\institution{Dalian University of Technology}
\city{Dalian}
\country{China}}
\email{zxt.dut@hotmail.com}

\author{Changya Li}
\affiliation{
  \institution{Dalian University of Technology}
  \city{Dalian}
  \country{China}}
\email{lichangya.dut@gmail.com}

\author{Feng Zhang}
\affiliation{
  \institution{Peking University}
  \city{Beijing}
  \country{China}}
\email{zfeng.maria@gmail.com}

\author{Wei Wang}
\affiliation{
  \institution{Shenzhen MSU-BIT University}
  \city{Shenzhen}
  \country{China}}
\email{ehomewang@ieee.org}

\author{Fenglong Ma}
\affiliation{
  \institution{The Pennsylvania State University}
  \city{University Park}
  \country{United States}}
\email{fenglong@psu.edu}

\author{Hong Yu}
\affiliation{
  \institution{Dalian University of Technology}
  \city{Dalian}
  \country{China}}
\email{hongyu@dlut.edu.cn}

\renewcommand{\shortauthors}{Han Liu et al.}

\begin{abstract}
Large language models (LLMs) have shown remarkable performance in various domains, but they are constrained by massive computational and storage costs. Quantization, an effective technique for compressing models to fit resource-limited devices while preserving generative quality, encompasses two primary methods: quantization aware training (QAT) and post-training quantization (PTQ). QAT involves additional retraining or fine-tuning, thus inevitably resulting in high training cost and making it unsuitable for LLMs. Consequently, PTQ has become the research hotspot in recent quantization methods. However, existing PTQ methods usually rely on various complex computation procedures and suffer from considerable performance degradation under low-bit quantization settings. To alleviate the above issues, we propose a simple and effective post-training quantization paradigm for LLMs, named SEPTQ. Specifically, SEPTQ first calculates the importance score for each element in the weight matrix and determines the quantization locations in a static global manner. Then it utilizes the mask matrix which represents the important locations to quantize and update the associated weights column-by-column until the appropriate quantized weight matrix is obtained. Compared with previous methods, SEPTQ simplifies the post-training quantization procedure into only two steps, and considers the effectiveness and efficiency simultaneously. Experimental results on various datasets across a suite of models ranging from millions to billions in different quantization bit-levels demonstrate that SEPTQ significantly outperforms other strong baselines, especially in low-bit quantization scenarios.
\end{abstract}


\maketitle

\section{Introduction}
\label{submission}
Large language models (LLMs) \cite{workshop2022bloom,radford2019language,liu2019roberta,devlin2018bert}, based on the transformer architecture \cite{vaswani2017attention}, have demonstrated exceptional performance across various downstream tasks. Despite their impressive capabilities, the immense number of parameters in these models presents significant deployment challenges on end-user devices due to high computational and storage demands. For instance, the GPT-175B model \cite{brown2020language}, which contains 175 billion parameters, requires approximately 350GB of storage space and more than 5 A100 GPUs (each with 80GB of memory) for inference. This underscores the urgent necessity for effective compression strategies for large language models. 

\begin{figure}[t]
    \centering
    \includegraphics[width=0.85\linewidth]{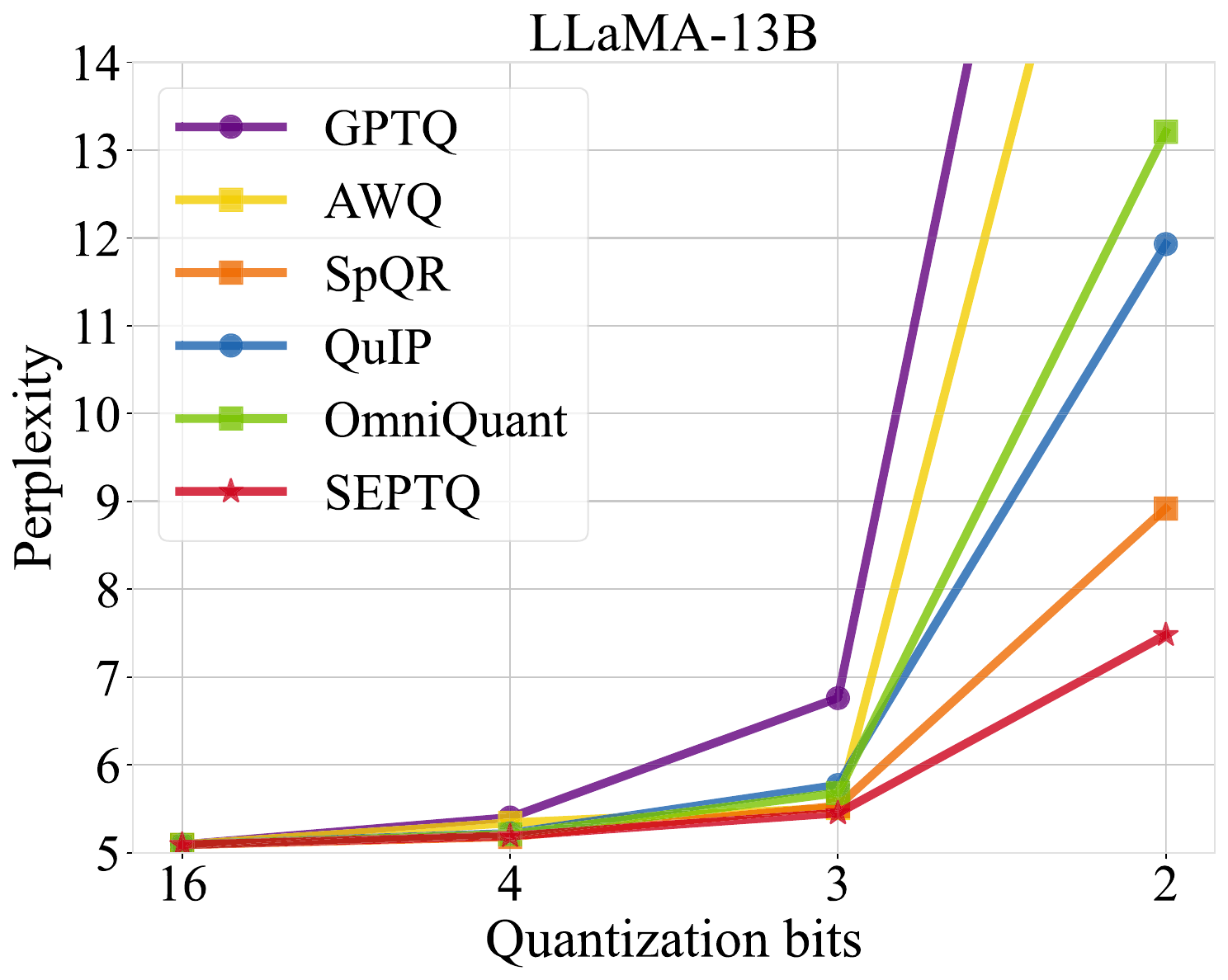}
    \caption{The perplexity results of widely-used quantization methods on LLaMA-13B, reported on WikiText2 (the lower value is better). The perplexity values of GPTQ and AWQ at 2-bit level are too large to display. All the approaches suffer from significant performance degradation when quantizing models to 2-bit level, and our proposed SEPTQ outperforms other baselines obviously.} 
    \label{over}
\end{figure}

Quantization has become a widely embraced technique to alleviate the storage and computational overhead of deep learning models \cite{DBLP:conf/icml/LeeKKL23,banner2019post,xiao2023smoothquant,DBLP:conf/icml/LiuLC23}, and its core idea is to convert model parameters from floating-point to lower-bit integer representation. In general, existing quantization methods can be divided into two categories: quantization aware training (QAT) and post-training quantization (PTQ). For QAT, it integrates the quantization objective into the model training process seamlessly, thus requiring additional retraining or fine-tuning. Its substantial training expense renders it impractical for LLMs. For PTQ, it diminishes the storage and computational complexity of the models without necessitating modifications to the LLM architecture or requiring a retraining process. Its simplicity and efficiency make it a common choice in existing quantization methods for LLMs.

Recently, several methods have been proposed for post-training quantization \cite{CBQ,RPTQ,liu2021post}. In particular, GPTQ \cite{frantar2022gptq} is a novel layer-wise quantization method based on approximate second-order information which aims to solve the problem of minimizing the layer-wise squared error. It can compress models to 3 bits or 4 bits without significant loss of accuracy. AWQ~\cite{AWQ} is an activation-aware method by considering the significance of weight channels corresponding to larger activation magnitudes. SpQR~\cite{SPQR} identifies sensitive weights and store them in higher precision, while compressing all other weights to 3-4 bits. QuIP~\cite{Quip} introduces an adaptive rounding procedure which minimizes a quadratic proxy of the weight error and can guarantee the incoherence of the weight and Hessian matrices. It is the first method which can obtain stable results in the 2-bit level compression. These methods have demonstrated considerable efficacy in reducing both computational and memory overhead for LLMs. Nevertheless, they either rely on intricate optimization procedures or suffer from substantial performance degradation in scenarios involving low-bit quantization \cite{OneBit,BiLLM,BitNet}, and Figure \ref{over} shows an example that the efficacy of different PTQ methods rapidly diminishes when the quantization bit-width is extremely low.

In this paper, we propose a simple and effective paradigm for post-training quantization, named SEPTQ, which not only ensures the effectiveness of the quantization process but also maintains its efficiency. SEPTQ is comprised of two modules: determining the quantization location and quantizing the model weights. For the first component, we adopt a static global strategy to identify the important parameters for each weight matrix and then obtain the corresponding mask matrix which represents the quantization locations. For the second component, we utilize the mask matrix that serves as a guide to quantize and update the associate weights column-by-column, where the update formula is derived under the constraint of the mask matrix and is able to precisely compensate for the rounding error that are inevitable when weights are quantized. Extensive experiments on a broad range of datasets and a diverse set of models, spanning from modest millions to immense billions of parameters, have unequivocally revealed the superiority of SEPTQ over other formidable baselines. Notably, in the low-bit (2-bit) quantization scenarios, where the challenge of maintaining model accuracy is most acute, SEPTQ has demonstrated significant performance gains, thereby further validating its effectiveness and robustness in the domain of model quantization.

\begin{figure*}[t]
    \centering
    \begin{subfigure}{0.495\linewidth}
        \centering
        \includegraphics[width=0.98\linewidth]{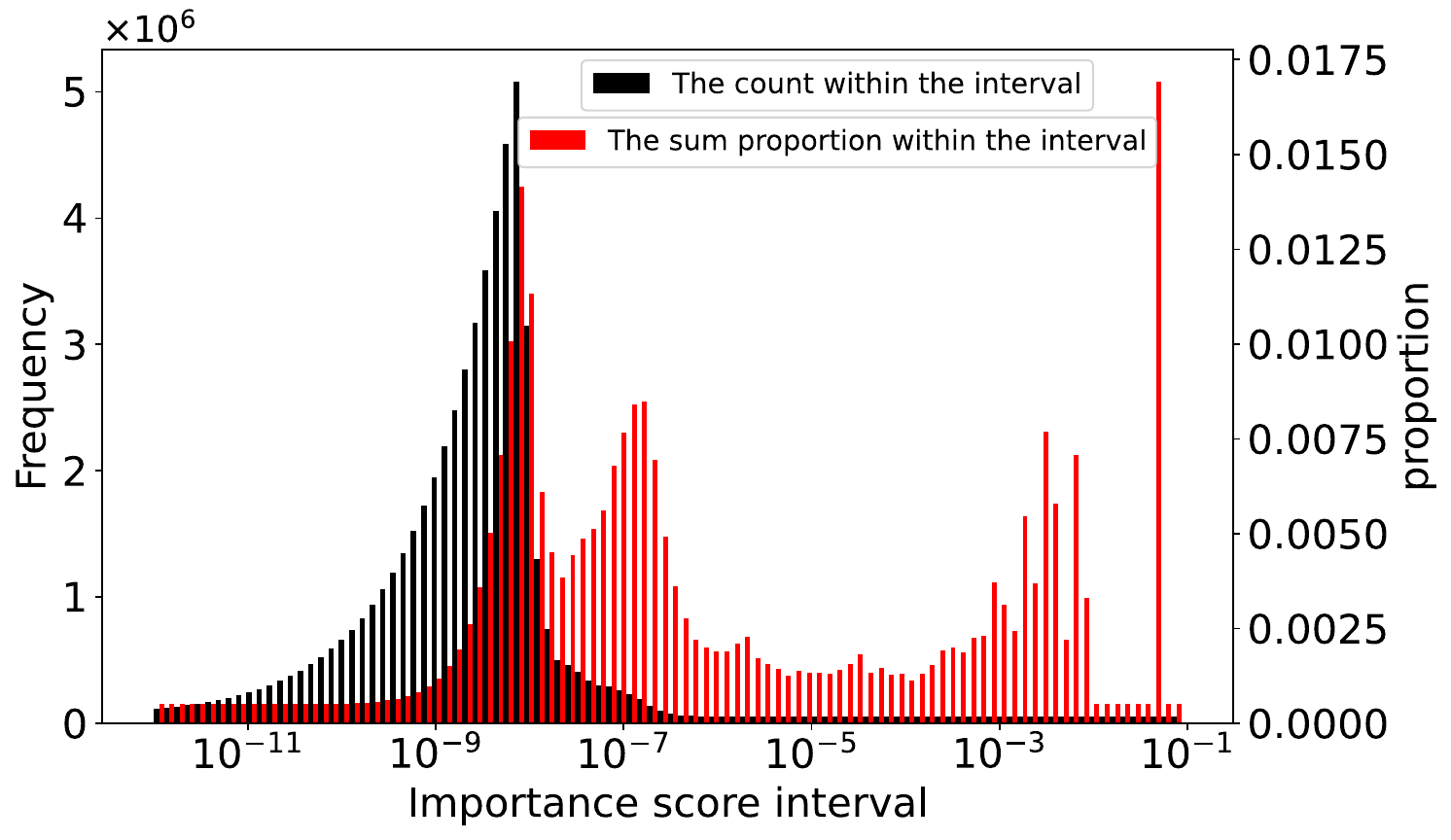}
        \caption{4-bit quantization}
    \end{subfigure}
    \hfill
    \begin{subfigure}{0.48\linewidth}
        \centering
        \includegraphics[width=0.98\linewidth]{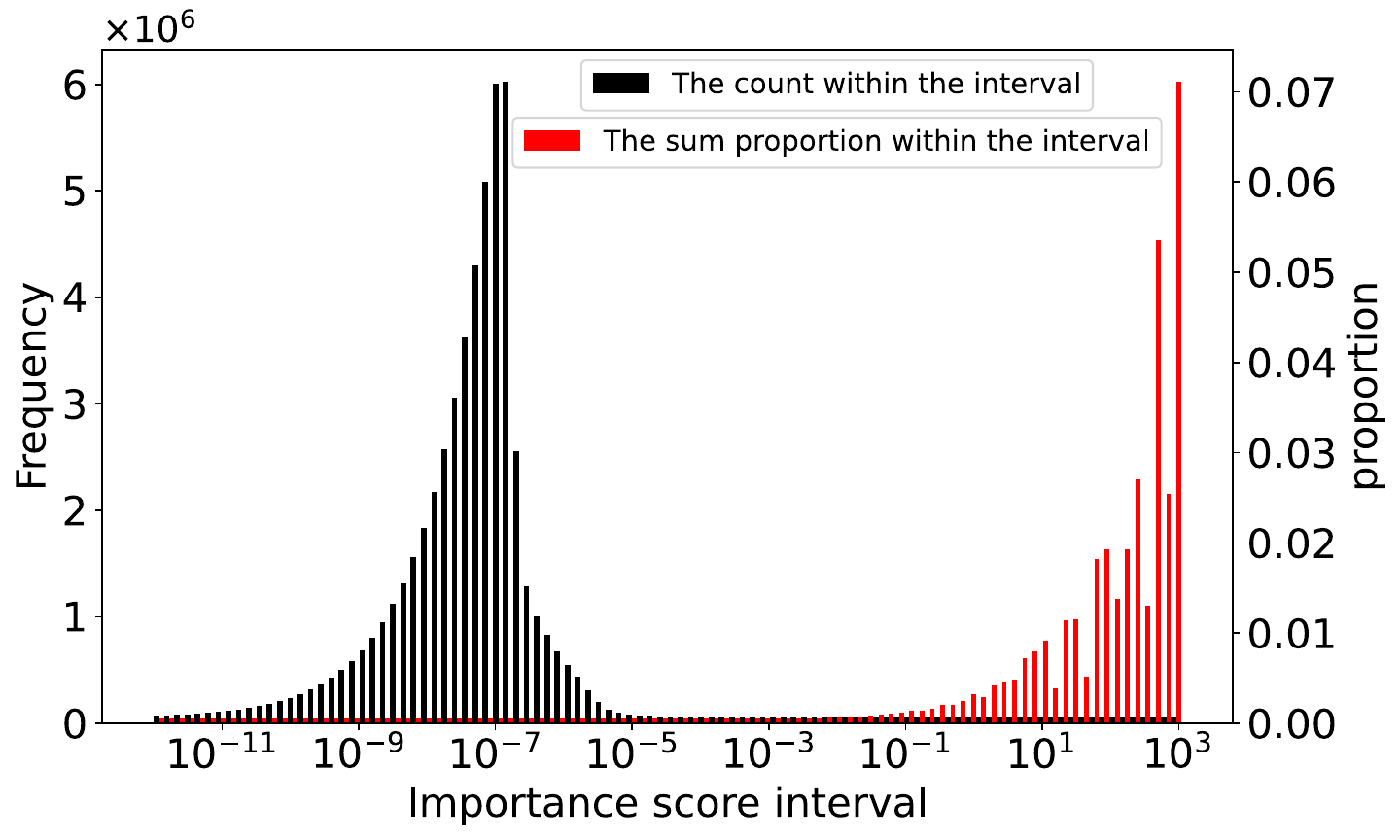}
        \caption{2-bit quantization}
    \end{subfigure}
    \caption{The frequency and proportion distributions of the importance scores in different quantization settings. In 4-bit quantization (a), a small number of weights have high importance scores and account for almost half of the importance score of the entire weight matrix. In 2-bit quantization (b), a small number of weights have extremely high importance scores and account for almost all of the importance score of the entire weight matrix.}
    \label{4biterror}
\end{figure*}

\section{Related Work}
Quantization aims to reduce the storage and computational overhead of deep learning models, which has become a widely accepted technique in model compression. Unlike traditional floating-point representation, quantization converts these numbers into integers or other discrete forms, thus significantly decreasing the storage requirement and computational complexity \cite{zongshu}. Existing model quantization methods can be divided into two main categories based on when quantization is applied: \textbf{Quantization Aware Training (QAT)} and \textbf{Post-Training Quantization (PTQ)}.

\subsection{Quantization Aware Training} QAT achieves the quantization during the model training \cite{tailor2020degree,kim2022understanding,ding20224}. LLM-QAT \cite{LLM-QAT} is a quantization-aware training approach that employs data-free distillation using data generated by a pre-trained model, bypassing data collection issues. OmniQuant~\cite{OmniQuant} achieves low-bit quantization by introducing learnable weight clipping (LWC) to optimize the clipping threshold and learnable equivalent transformation (LET) to shift the quantization challenge from activations to weights, effectively handling activation outliers. Although the QAT methods often have good performance, their processes are often tedious and time-consuming.

\subsection{Post-Training Quantization} PTQ quantizes the model after it has completed training \cite{fang2020post,li2021brecq,QMoE}, which aims to reduce the storage and computational complexity of LLMs without modifying the model architecture or requiring a retraining process. Compared with QAT, PTQ is simple and efficient in achieving model compression. Optimal brain quantization (OBQ) ~\cite{frantar2022optimal} extends the traditional second-order weight pruning framework of optimal brain surgeon (OBS) ~\cite{hassibi1992second,kurtic2022optimal}. It quantizes weights based on quantization errors and achieves good results on smaller models without retraining. GPTQ~\cite{frantar2022gptq} introduces a layer-wise quantization technology based on approximate second-order information. It quantizes the weights column-wise and updates the remaining weights according to the OBQ method until all weights are quantized. AWQ~\cite{AWQ} looks for optimal per-channel scaling factors to protect these outlier (important) weights by looking at activation patterns rather than the weights themselves. However, if the scaling factor is too large, it will increase the quantization loss of non-outlier weights, while if the scaling factor is too small, it cannot protect these outlier weights well. SpQR~\cite{SPQR} achieves near-lossless compression at different model sizes by identifying and isolating those outlier weights that cause particularly large quantization errors and storing them with higher precision. However, it uses the dynamic local strategy to find the outlier weights which is time-consuming and suboptimal. QuIP~\cite{Quip} employs an adaptive rounding procedure to minimize a quadratic approximation target and then performs efficient pre- post-processing to ensure the incoherence of the weight and Hessian matrices by multiplying them by a Kronecker product of random orthogonal matrices. It is a powerful method and achieves 2-bit level quantization, but its computation cost is a little expensive. 

\section{Problem Formulation}
In this paper, we focus on the problem of post-training quantization \cite{dettmers2022llm,nagel2020up,DBLP:conf/icml/HubaraNHBS21}, and aims to design a simple and effective post-training quantization framework for large language models. Specifically, let $\mathbf{W} \in \mathbb{R}^{d_{row} \times d_{col}} $ denote the weight matrix of a linear layer and $\mathbf{X} \in \mathbb{R}^{d_{col} \times n}$ represent an input matrix, where $n$ is the number of samples. The whole goal is to discover a compressed version of $\mathbf{W}$ (denoted as $\mathbf{\widehat{W}}$), which can minimize the squared error associated with the full precision layer output. Formally, 
\begin{equation}
\label{OBC}
\text{argmin}_{\mathbf{\widehat{W}}}\|\mathbf{W}\mathbf{X} -\widehat{\mathbf{W}}\mathbf{X}  \|_\text{F}^2,~~~~~\ \ \ \ \ \text{s.t.}\ \ \widehat{\mathbf{W}}=\text{quant}(\mathbf{W}),
\end{equation}
where $\|\cdot\|_\text{F}$ is the Frobenius norm, and $\text{quant}(\cdot)$ refers to the round-to-nearest (RTN) method  \cite{yao2022zeroquant}, which can round a model weight to the nearest value on the quantization grid. In particular, if $\mathbf{W}_{i,j}$ denotes the $i$-th row and $j$-th element of $\mathbf{W}$, then $\mathbf{\widehat{W}}_{i,j}$ can be calculated as:
\begin{equation}
\label{RTN}
\mathbf{\widehat{W}}_{i,j}=S (\mathbf{{W}}^q_{i,j}-Z),\ \  
\mathbf{{W}}^q_{i,j}=\text{clip} ( \lfloor\frac{\mathbf{W}_{i,j}}{S} \rceil  + Z, 0, 2^N-1 ),
\end{equation}
where $S$ is the quantization scale parameter, $Z$ is the zero point parameter, $N$ is the number of quantization bits, $\lfloor\frac{\mathbf{W}_{i,j}}{S} \rceil$ is the rounding function, and $\text{clip}(\cdot)$ is the truncation function which can guarantee that the value is the range of 0 to $2^N-1$.

\section{The Proposed Paradigm}
The proposed quantization paradigm consists of two components: determining the quantization location and quantizing the model weights. In this section, we will introduce these modules in details.

\subsection{Determining the Quantization Location}
In general, different weight parameters in a neural network contribute unequally \cite{SPQR}. Therefore, for the quantization task, it is a crucial step to determine the quantization location (weight parameter). Intuitively, if quantizing a weight parameter leads to a large rounding error, the weight parameter can be treated as an important parameter which tends to be reserved. In contrast, if quantizing a weight parameter affects the rounding error sightly, the weight parameter can be treated as an unimportant parameter which should be quantized. 

Furthermore, if we utilize the square error between $\mathbf{W}\mathbf{X}$ and $\widehat{\mathbf{W}}\mathbf{X}$ to obtain the rounding error, we can have the following formula to calculate the importance score $s_{i,j}$ for each element $\mathbf{W}_{i,j}$ in $\mathbf{W}$:
\begin{equation}
\label{Eq3}
\begin{split}
s_{i,j} &= \text{min}_{\widehat{\mathbf{W}}(i,j)} \left\| \mathbf{W}\mathbf{X} - \widehat{\mathbf{W}}(i,j)\mathbf{X} \right\|_\text{F}^2 \\
&\text{s.t.} \ \ \widehat{\mathbf{W}}(i,j) = \text{quant}(\mathbf{W}(i,j)),
\end{split}
\end{equation}
where $\widehat{\mathbf{W}}(i,j)$ denotes the matrix $\mathbf{W}$ with only $\mathbf{W}_{i,j}$ is quantized and unquantized elements in the $i$-th row are updated. By leveraging Taylor formula and the Lagrangian multiplier method (The detailed proof procedure can refer to Appendix \ref{APPA}), the importance score $s_{i,j}$ can be derived as:
\begin{equation}
\label{Eq4}
s_{i,j}=\frac{\left(\mathbf{W}_{i,j} - \text{quant}(\mathbf{W}_{i,j})\right)^2}{2[\mathbf{XX}^T]^{-1}_{j,j}}.
\end{equation}

By utilizing Eq.~(\ref{Eq4}), in terms of a particular input matrix $\mathbf{X}$, we can obtain the importance score of each element in $\mathbf{W}$. Following that, we sort all the importance scores in the descending order, and select the top $p\%$ elements as reserved weight parameters, and the remaining $1-p\%$ elements as quantized weight parameters. In this paper, we usually set the ratio $p$ to be less than or equal to $1\%$. 

In addition, we further analyze the distribution of important scores. Specifically, we take the output projection matrix of the fourth layer in the OPT-30B model as the example to explore the distribution of important scores (For more experimental results, please refer to the Appendix \ref{APPD}). We first calculate the importance score for each weight of the output projection matrix, and then draw the frequency distribution histogram and proportion histogram of these scores, which are shown in Figure \ref{4biterror}. The black bars represent the number of weights within the importance score interval, and the red bars represent the proportion of the sum of the importance scores within the importance score interval to the sum of all importance scores.

A salient feature observed from Figure \ref{4biterror} (a) is the prevalence of lower importance scores, with a minority exhibiting higher values. Despite their scarcity, these high-scoring weights exert a disproportionately large influence on the total importance score. The analysis extends to Figure \ref{4biterror} (b), revealing that the vast majority of importance scores fall below $10^{-7}$. Interestingly, despite their numerical abundance, these scores contribute minimally to the overall sum, highlighting a more pronounced long-tail effect in 2-bit quantization compared to its 4-bit counterpart. This exacerbation of the long-tail phenomenon underscores the critical role of a select few high-importance weights in the quantization process, particularly as the quantization bit-width diminishes.

The implications of these findings are twofold. Firstly, the presence of a few high-importance weights indicates that their impact on the quantization process is non-trivial. Secondly, as quantization bit-width decreases, the influence of these critical weights becomes more pronounced. Effective weight selection is crucial for maintaining model performance in quantization. 
Based on these insights, by preserving the quantization of important weights and selectively quantizing the less critical ones, the precision of compression can be enhanced. Moreover, the benefits of this strategy increase as the quantization bit width decreases.

For the ease of representing the quantization locations, we introduce the mask matrix $\mathbf{M}$ to express whether an element in the weight matrix $\mathbf{W}$ is selected as the quantization location. If $\mathbf{M}_{i,j}=0$, it indicates $\mathbf{W}_{i,j}$ needs to be quantized. And if $\mathbf{M}_{i,j}=1$, it indicates $\mathbf{W}_{i,j}$ needs to be reserved.

\subsubsection{Static Strategy vs. Dynamic Strategy} By static strategy, we mean that the importance scores of each weight matrix are calculated at once. By dynamic strategy, we mean that the importance scores of each weight matrix are calculated sequentially (column-by-column), as previous works like GPTQ \cite{frantar2022gptq} update some corresponding weight parameters to compensate for rounding errors after quantizing each column. To analyze which strategy is more appropriate, we attempt to visualize the determined quantization locations (the mask matrix $\mathbf{M}$) by static strategy and dynamic strategy. Specifically, we take the $7$-th layer weight matrix of the OPT-30B model \cite{zhang2022opt} as example and set the $p=1\%$. As the layer dimension is too large, we divide the whole matrix into different blocks with the size of 128 rows $\times$ 128 columns, and sum the element values of each block in $\mathbf{M}$. Figure~\ref{fig2} shows the the determined quantization location distribution. It can seen that the overall distribution situation is very similar, which indicates that no matter using static strategy or dynamic strategy to determine the quantization locations, the performance will be stable. The dynamic strategy will inevitably add extra computation cost which is caused by updating weight parameters. Therefore, we employ the static strategy to determine the quantization locations.

\subsubsection{Global Strategy vs. Local Strategy} 
\begin{figure}
    \includegraphics[width=1\linewidth]{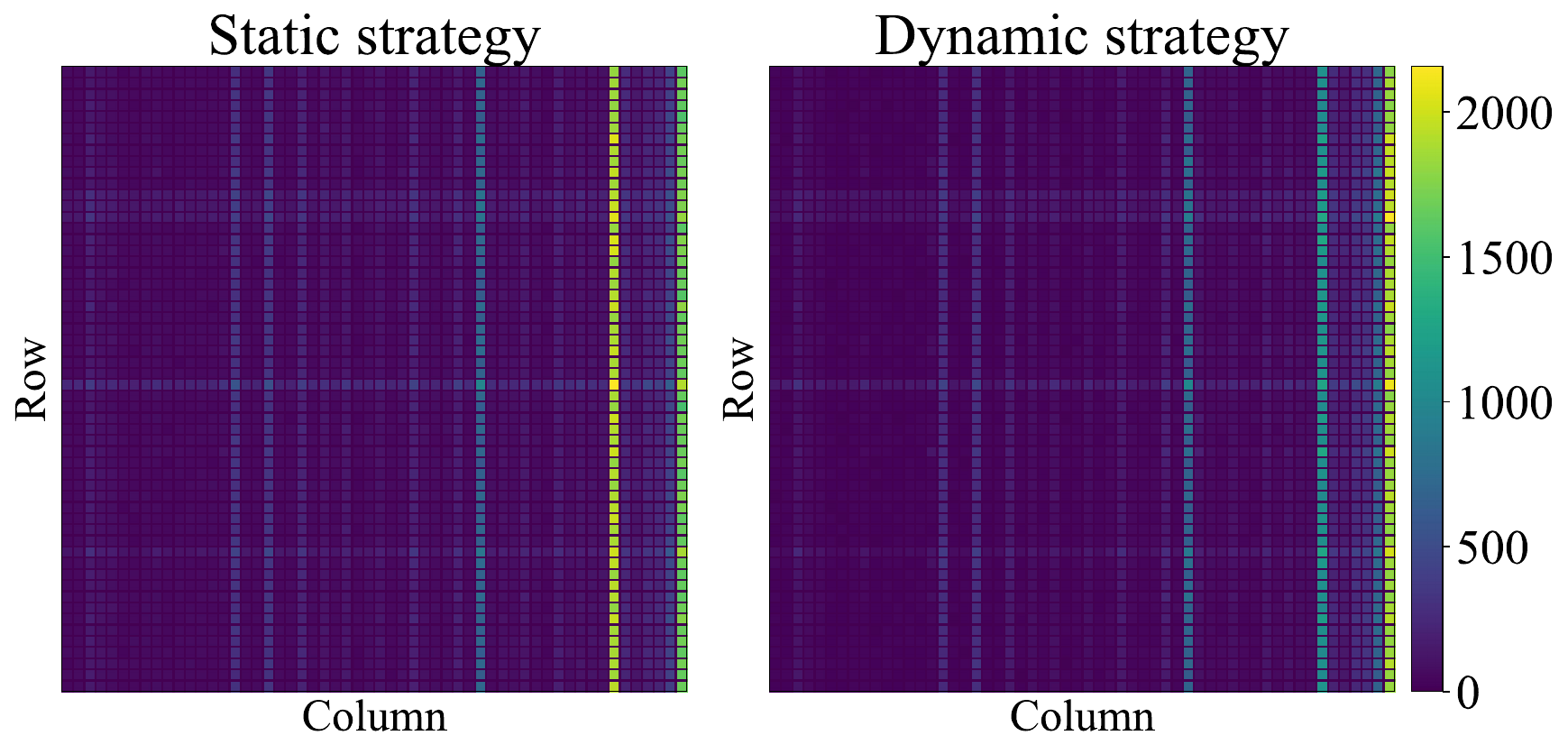}
    \caption{The visualization of determined quantization location distribution by static strategy (ours) and dynamic strategy (GPTQ). The value of each block is the sum of the values in the corresponding locations of the mask matrix $\mathbf{M}$.}
    \label{fig2}
\end{figure}

By global strategy, we mean that selecting the top $p\%$ elements from the whole matrix as reserved weight parameters. By local strategy, we mean that selecting the top $p\%$ elements from each block as reserved weight parameters. As illustrated in Figure~\ref{fig2}, we can get that the important weight parameters tend to cluster in some specific rows or columns. That is to say, some blocks clearly contain a large amount of important weight parameters, while some blocks do not have any important weight parameters which are worth preserving. Based on the above observation, we utilize the global strategy to determine the quantization locations.

\subsection{Quantizing the Model Weights}
\begin{figure*}[t]
    \centering
    \includegraphics[width=0.95\linewidth]{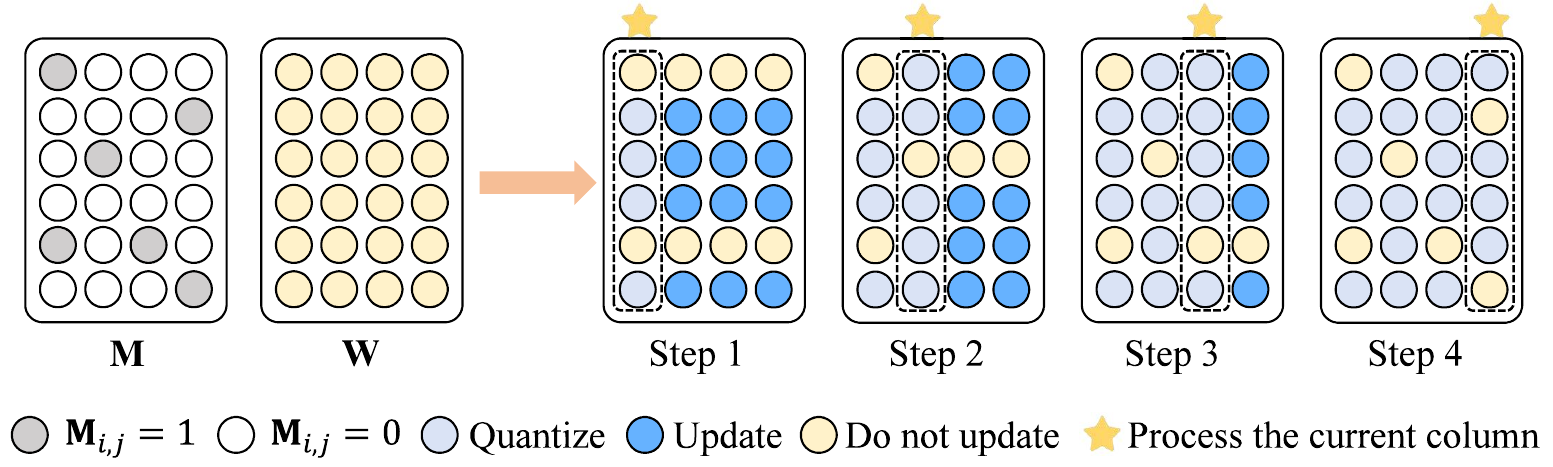}
    \caption{The visualization of quantizing the model weights. Given the original weight matrix $\mathbf{W}$ and the mask matrix $\mathbf{M}$, we process the matrix $\mathbf{W}$ in a column-by-column manner. If $\mathbf{M}_{i,j}=1$, the corresponding weights and the subsequent weights in the same row both remain unchanged (light yellow). If $\mathbf{M}_{i,j}=0$, the corresponding weights are quantized (light blue) and the subsequent weights in the same row are updated to compensate for the quantization loss (dark blue).}
    \label{QPA}
\end{figure*}

This procedure consists of two steps: quantizing the model weights in selected quantization locations and updating the model weights in the non-quantization locations. 

In terms of quantizing the model weights in selected quantization locations, as the computation complexity of directly solving Eq. (\ref{OBC}) is excessively high, inspired by \cite{frantar2022gptq}, we quantize the model weights in a column-by-column manner and the specific quantization computation formula can refer to Eq. (\ref{RTN}).

In terms of updating the model weights in the non-quantization locations, we calculate the increment for the non-quantization weights to compensate the rounding error caused by quantized weights. In particular, after obtaining the mask matrix which can represent the quantization locations, we can modify the constraint in Eq. (\ref{OBC}) as:
\begin{equation}
    \label{Eq5}
   \widehat{\mathbf{W}}= \text{quant}(\mathbf{W}) + \mathbf{M} \odot \left(\mathbf{W} - \text{quant}(\mathbf {W})\right).
\end{equation}

According to Eq. (\ref{Eq5}), it is easy to find that if $\mathbf{M}_{i,j}=0$, $\widehat{\mathbf{W}}_{i,j}= \text{quant}(\mathbf{W}_{i,j})$. And if $\mathbf{M}_{i,j}=1$, $\widehat{\mathbf{W}}_{i,j}= \mathbf{W}_{i,j}$. Furthermore, we employ the similar method in \cite{frantar2022optimal} to solve the Eq. (\ref{OBC}) with the constraint in Eq. (\ref{Eq5}) (The detailed proof procedure can refer to Appendix \ref{APPB}).

Specifically, assume that when processing the $j$-th column in $\mathbf{W}$, the $i$-th element of this column is quantized, then we need to update the corresponding weights in the $i$-th row and greater than the $j$-th column of $\mathbf{W}$. we let $\delta=(\mathbf{\widehat{W}}_{i,:}-\mathbf{W}_{i,:})^{\top}$ denote the increment of the weight of the i-th row, according to Eq. (\ref{Eq5}), we can get the constraint condition of the i-th row as
\begin{equation}
    \mathbf{e}_j^T \boldsymbol{\delta}+(\mathbf{W}_{i,j}-\widehat{\mathbf{W}}_{i,j})=0,
\end{equation}
where $\mathbf{e}_j$ represents the unit vector with 1 in the $j$-th position. This problem has the following Lagrangian\cite{frantar2022optimal}:
\begin{equation}
L(\boldsymbol{\delta}, \lambda) = \frac{1}{2} (\boldsymbol{\delta}^T \mathbf{H}  \boldsymbol{\delta}) + \lambda (\mathbf{e}_j^T  \boldsymbol{\delta} + (\mathbf{W}_{i,j} - \widehat{\mathbf{W}}_{i,j})).
\end{equation}
By solving the derivatives of $\boldsymbol{\delta}$ and $\lambda$, we can get:
\begin{equation}
\label{Eq6}
    \boldsymbol{\delta} = -\frac{\mathbf{W}_{i,j} - \widehat{\mathbf{W}}_{i,j}}{[\mathbf{H}^{-1}]_{jj}} [\mathbf{H}^{-1}]_{:,j},
\end{equation}
where $\mathbf{H}=2 \mathbf{X}\mathbf{X}^T$, and we only need take greater than the $j$-th elements of $\boldsymbol{\delta}$ to update the associated weights. To illustrate the above procedure clearly, we take an example to visualize the compression pipeline of our algorithm in Figure \ref{QPA}.

\begin{algorithm}[!t]
\caption{The SEPTQ Algorithm}
\label{algorithm1}
\parbox[t]{\linewidth}{\textbf{Input}: The weight matrix $\textbf{W}$, the calibration dataset $\textbf{X}$, the ratio parameter $p$, the blocksize $B$}\\
\parbox[t]{\linewidth}{\textbf{Output}: The quantized weight matrix $\widehat{\mathbf{W}}$}\\
\begin{algorithmic}[1]
   \STATE $\mathbf{Q} \leftarrow \textbf{0}_{d_\text{row} \times d_\text{col}}$ \hfill // Save the quantized weights
   \STATE $\textbf{E} \leftarrow \textbf{0}_{d_{\text{row}} \times B}$ 
   \hfill // Initialize the block quantization errors
   \STATE $\textbf{M} \leftarrow \textbf{1}_{d_{\text{row}} \times d_\text{col}}$
   \hfill // Initialize the static global mask matrix
   \STATE$\mathbf{H}^{-1} \leftarrow (2\mathbf{XX}^T+\lambda \mathbf{I})^{-1}$
   \STATE$\mathbf{H}^{-1} \leftarrow \text{Cholesky}(\mathbf{H}^{-1}) $ \hfill //Use Cholesky decomposition
   \STATE$\textbf{W}^Q \leftarrow \text{quant}(\mathbf{W}) $ \hfill 
   \STATE $\textbf{M} \leftarrow \text{mask of } p\% \ \text{weights}$ by Eq. (\ref{Eq4}) \hfill 
   \FOR{$i = 0,B,2B,\dots$}
   \FOR{$j=i,\dots,i+B-1$}
      \STATE $\textbf{Q}_{:,j} \leftarrow \text{quant}(\textbf{W}_{:,j})$
      \STATE $\textbf{Q}_{:,j} \leftarrow \textbf{Q}_{:,j} + \textbf{M}_{:,j}\odot(\textbf{W}_{:,j}-\textbf{Q}_{:,j})$ \hfill // Calculate by Eq. (\ref{Eq5})
      \STATE $\textbf{E}_{:,j-i} \leftarrow \left(\textbf{W}_{:,j} -\textbf{Q}_{:,j}\right)/[\textbf{H}^{-1}]_{jj}$\hfill  // Quantization error
      \STATE $\textbf{W}_{:,j:(i+B)} \leftarrow \textbf{W}_{:,j:(i+B)} - \textbf{E}_{:,j-i}  \textbf{H}^{-1}_{j,j:(i+B)}$ \hfill  // Update the weights by Eq. (\ref{Eq6})
   \ENDFOR
   \STATE $\textbf{W}_{:,(i+B):} \leftarrow \textbf{W}_{:,(i+B):} - \textbf{E}  \textbf{H}^{-1}_{i:(i+B),(i+B):}$
   \ENDFOR
   \STATE $\widehat{\mathbf{W}} \leftarrow \mathbf{Q}$ 
\end{algorithmic}
\end{algorithm}

\subsection{The Overall Algorithm}
The detailed algorithm procedure is shown in Algorithm \ref{algorithm1}. We implement our algorithm based on the GPTQ framework \cite{frantar2022gptq}. Compression is performed per linear layer. Specifically, given a weight matrix $\mathbf{W}$ and a calilbration dataset $\mathbf{X}$, SEPTQ first computes the Hessian matrix and the inverse of the Hessian matrix approximated through Cholesky decomposition. Then it calculates the mask $\mathbf{M}$ in a static and global manner. Subsequently, some associated weights are quantized and updated. During the update process, we apply our algorithm to each block based on the lazy batch-update, and perform the global update after a block is processed \cite{frantar2022gptq}. Finally, it can obtain the quantized weight matrix $\widehat{\mathbf{W}}$.

\section{Experiments}
\begin{table*}[t]
\centering
\caption{The perplexity results of the OPT model family on the C4 dataset (The lower value is better). ``Full''means the full precision method (the original model). For SEPTQ, it is exactly 2.1-bit, 3.1-bit and 4.1-bit respectively.}
\label{table1}
\renewcommand{\arraystretch}{1.05}
\tabcolsep=0.4cm
\begin{tabular}{l|c|c|c|c|c|c|c|c|c} 
\toprule
Method  & Bits     & 125M   & 350M   & 1.3B  & 2.7B   & 6.7B  & 13B    & 30B    & 66B   \\
\midrule
Full  & 16          &26.56&	22.59&	16.07&	14.34&	12.71&	12.06&	11.44&	10.99  \\
\midrule
GPTQ & \multirow{3}{*}{2}       &2381.23&	6329.39&	4342.76&	3915.16&	522.25&	125.88&	29.06&	253.49  \\
QuIP  &       &177.40&	320.00	&29.78&	38.07&	21.62&	16.60&	13.55&	17.64  \\
\textbf{SEPTQ} &   &\textbf{53.75}&	\textbf{46.33}&	\textbf{21.64}&	\textbf{18.18}&	\textbf{15.19}&	\textbf{13.97}&	\textbf{12.80}&	\textbf{11.96} \\
\midrule
GPTQ & \multirow{3}{*}{3}    &42.01&	31.33&	21.65&	18.20&	17.23&	13.36&	12.22&	13.79 \\
QuIP &        &30.92&	25.48&	17.12&	15.63&	13.30&	12.39&	11.66&  11.19  \\
\textbf{SEPTQ}&   &\textbf{30.37}&	\textbf{25.18}&	\textbf{16.99}&	\textbf{14.97}&	\textbf{13.05}&	\textbf{12.32}&	\textbf{11.61}&	\textbf{11.11}  \\
\midrule
GPTQ  &  \multirow{3}{*}{4}    &29.33&	24.62&	16.98&	15.01&	13.18&	12.26&	11.56&	11.25  \\
QuIP  &      &\textbf{27.63}&	\textbf{23.23}&	16.38&	\textbf{14.55}&	12.86&	12.16&	11.50&11.03	 \\
\textbf{SEPTQ} &    &27.78&	\textbf{23.23}&	\textbf{16.37}&	\textbf{14.55}&	\textbf{12.82}&	\textbf{12.13}&	\textbf{11.49}&	\textbf{11.02}\\
\bottomrule
\end{tabular}
\end{table*}

\begin{table*}[!t]
\centering
\caption{The perplexity results of the OPT model family on the WikiText2 dataset (The lower value is better). ``Full''means the full precision method (the original model). For SEPTQ, it is exactly 2.1-bit, 3.1-bit and 4.1-bit respectively.}
\label{table2}
\renewcommand{\arraystretch}{1.05}
\tabcolsep=0.4cm
\begin{tabular}{l|c|c|c|c|c|c|c|c|c} 
\toprule
Method  & Bits     & 125M   & 350M   & 1.3B  & 2.7B   & 6.7B  & 13B    & 30B    & 66B   \\
\midrule
Full      & 16         & 27.65  & 22.00  & 14.63 & 12.47  & 10.86 & 10.13  & 9.56   & 9.34  \\
\midrule
GPTQ  & \multirow{3}{*}{2}     & 4761.03  & 17558.09  & 8293.39 & 8575.29  & 3110.27 & 303.77    & 54.52  & 619.32   \\
 QuIP &        & 347.40  & 672.30  & 41.64 & 2998.00  & 22.33 & 16.02  & 11.48  & 13.86    \\
\textbf{SEPTQ}&   & \textbf{68.62}  & \textbf{54.99}  & \textbf{20.11} & \textbf{16.20}  & \textbf{12.97} & \textbf{11.82 } & \textbf{10.75}  & \textbf{9.93 } \\

\midrule
GPTQ   & \multirow{3}{*}{3}      & 52.94  & 33.85  & 21.82 & 16.95~ & 14.94 & 11.80  & 10.35  & 13.69 \\
QuIP  &      & 34.22  & 25.19  & 16.21 & 17.44  & 11.51 & 10.50   & 9.79   & 9.41     \\
\textbf{SEPTQ}&   & \textbf{33.77}  & \textbf{24.83}  & \textbf{15.46} & \textbf{12.74}  & \textbf{11.14} & \textbf{10.33}  & \textbf{9.61}   & \textbf{9.27}  \\
\midrule
GPTQ     & \multirow{3}{*}{4}       & 31.31  & 24.03  & 15.50 & 12.84~ & 11.37 & 10.31  & 9.61   & 9.42  \\
QuIP &        & 33.35  & \textbf{22.50}  & \textbf{14.88} & \textbf{12.39}  & \textbf{10.98} & \textbf{10.21}   & 9.60   & 9.42     \\
\textbf{SEPTQ} &    & \textbf{29.00}  & 22.78  & \textbf{14.88} & 12.46  & \textbf{10.98} & 10.22  & \textbf{9.54}   & \textbf{9.27}\\
\bottomrule
\end{tabular}
\end{table*}

\begin{table*}[!t]
\centering
\caption{The perplexity results of the OPT model family on the PTB dataset (The lower value is better). ``Full''means the full precision method (the original model). For SEPTQ, it is exactly 2.1-bit, 3.1-bit and 4.1-bit respectively.}
\label{table3}
\renewcommand{\arraystretch}{1.05}
\tabcolsep=0.4cm
\begin{tabular}{l|c|c|c|c|c|c|c|c|c} 
\toprule
Method  & Bits     & 125M   & 350M   & 1.3B  & 2.7B   & 6.7B  & 13B    & 30B    & 66B   \\
\midrule
Full      & 16          & 38.99  & 31.07	 &20.29	 &17.97	 &15.77	  &14.52   &14.04	&13.36  \\
\midrule
GPTQ & \multirow{3}{*}{2}       &4350.98	&11670.97	&7472.98	&6787.27	&3195.70	&246.22	&93.02	&630.42  \\
QuIP  &      &430.00	&744.20	&47.72	&63.59	&31.73	&21.64	&17.40	&17.50    \\
\textbf{SEPTQ} &  &\textbf{96.09}	&\textbf{74.70}	&\textbf{31.23}	&\textbf{24.69}	&\textbf{19.28}	&\textbf{17.88}	&\textbf{16.40}	&\textbf{14.93} \\
\midrule
GPTQ & \multirow{3}{*}{3}     &73.58	&45.95	&30.85	&24.76	&22.31	&16.74	&15.38	&26.46 \\
QuIP   &       &\textbf{47.34}	&\textbf{35.65}	&22.76	&20.79	&16.52	&15.05	&14.37	&13.55   \\
\textbf{SEPTQ}&   &48.21	&36.11	&\textbf{22.24}	&\textbf{18.89}	&\textbf{16.27}	&\textbf{14.88}	&\textbf{14.25}	&\textbf{13.51}  \\
\midrule
GPTQ  & \multirow{3}{*}{4}      & 45.60&34.15&22.04&19.15&16.49&14.85&14.26&13.81  \\
QuIP    &    &\textbf{40.80}&\textbf{32.57}&20.87&18.42&15.93&14.69&14.18&	\textbf{13.40}   \\
\textbf{SEPTQ}      &    &41.26	&32.65	&\textbf{20.78}&	\textbf{18.25}	&\textbf{15.87}	&\textbf{14.67}&	\textbf{14.12}&	\textbf{13.40}\\
\bottomrule
\end{tabular}
\end{table*}

\begin{figure*}[h]
    \centering
    \includegraphics[width=0.9\linewidth]{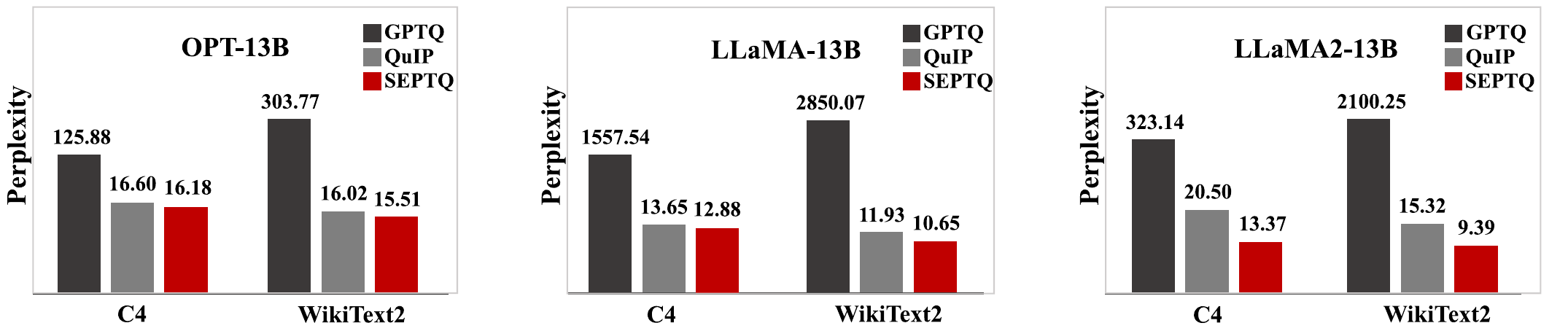}
    \caption{The perplexity results of the OPT-13B, LLaMA-13B and LLaMA2-13B models on C4 and WikiText2 datasets (The lower value is better). For GPTQ and QuIP, they are 2-bit models. And for SEPTQ, it is a 2.01-bit model.}
    \label{2bit}
\end{figure*}

\begin{figure*}[!t]
  \centering
  \begin{minipage}{0.48\textwidth}
    \centering
    \includegraphics[width=0.85\linewidth]{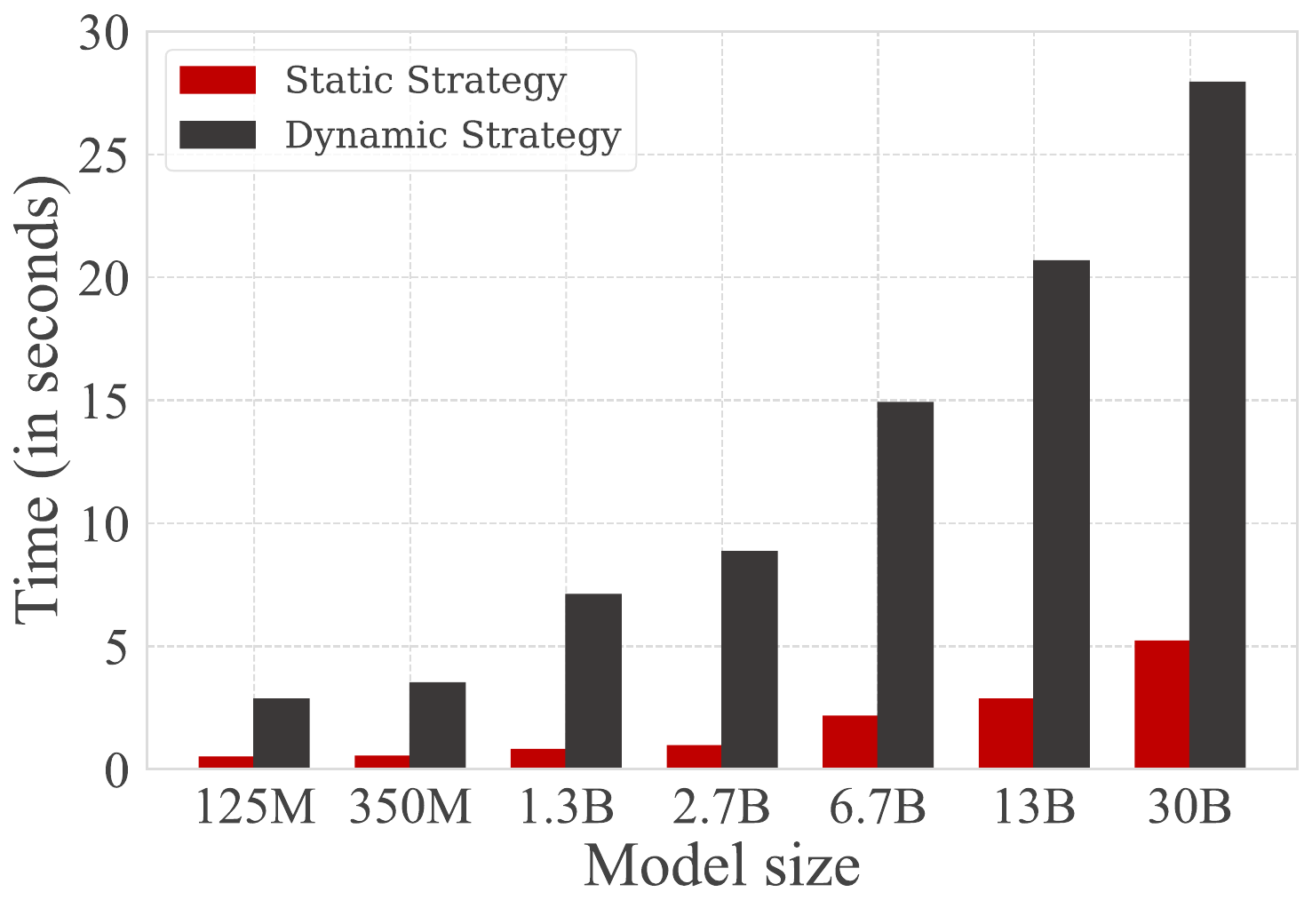}
    \caption{The efficiency results by using the static strategy and dynamic strategy respectively.}
    \label{fig:image1}
  \end{minipage}
  \hfill
  \begin{minipage}{0.48\textwidth}
    \centering
    \includegraphics[width=0.8\linewidth]{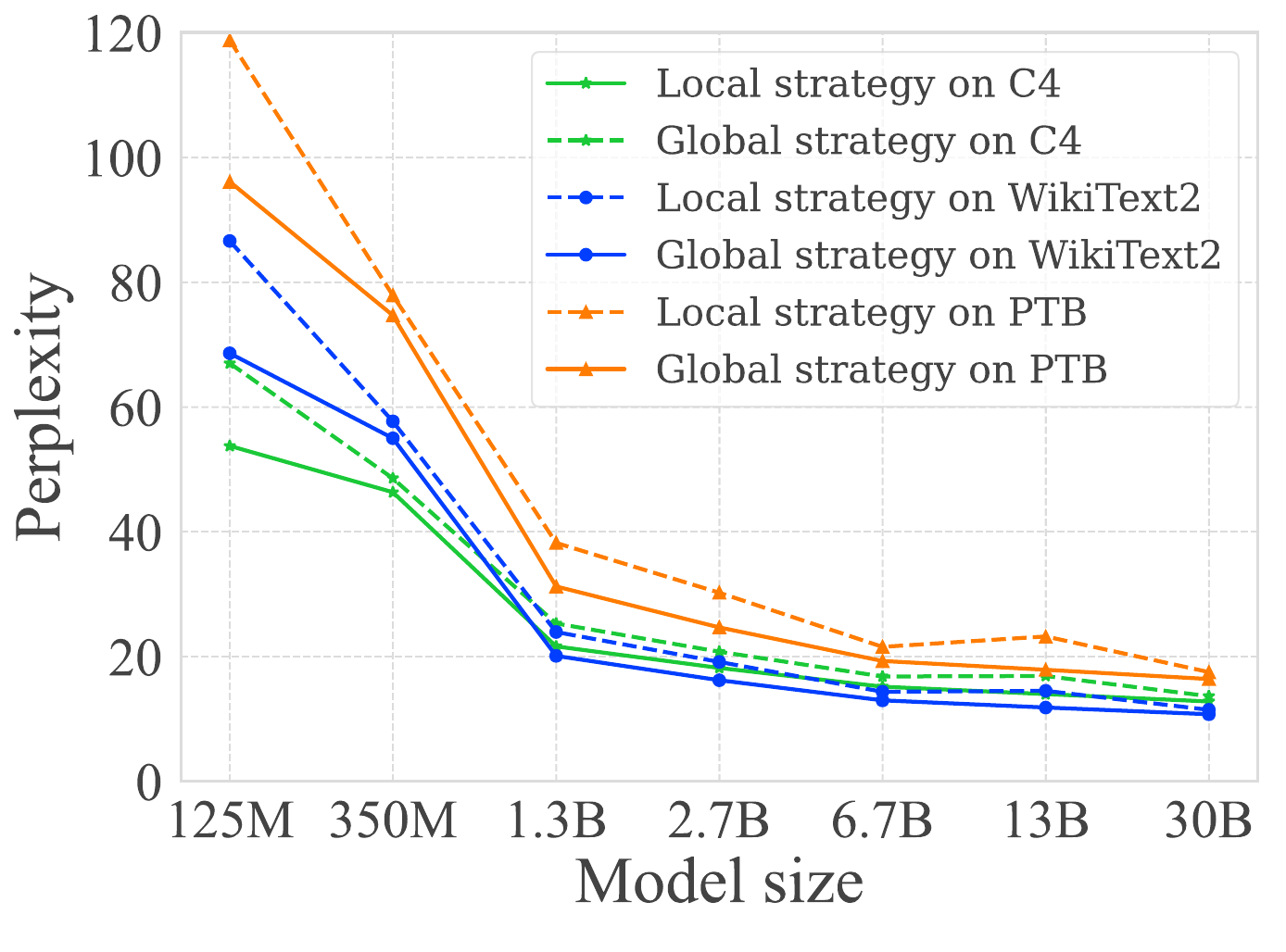}
    \caption{The effectiveness results by using the global strategy and local strategy respectively.}    
    \label{fig:image2}
  \end{minipage}
\end{figure*}

\subsection{Experimental Settings}
To validate the superiority of our proposed method, we follow \cite{frantar2022gptq} to conduct two types of experiments: perplexity experiments and accuracy experiments. Following \cite{frantar2022gptq,Quip}, as during the experimental procedure, each model does not see any task-specific data, and the accuracy experiments are also named as zero-shot experiments.

\subsubsection{Models and Datasets} For perplexity experiments, we evaluate different models by quantizing the OPT model family \cite{zhang2022opt} from 125 millions to 66 billions on C4 \cite{raffel2020exploring}, WikiText2 \cite{merity2016pointer} and PTB \cite{marcus1994penn} datasets. For zero-shot experiments, we measure different models by quantizing the popular LLaMA model \cite{touvron2023llama} from 7 billions to 30 billions on  PIQA \cite{PIQA}, ARC-easy and ARC-challenge \cite{clark2018think}, HellaSwag \cite{HellaSwag} and WinoGrande \cite{sakaguchi2021winogrande} datasets. For the calibration dataset $\mathbf{X}$, we follow \cite{frantar2022gptq} to use 128 random 2048 token segments from the C4 dataset. 

\subsubsection{Baselines} We compare our method with the strong post-training quantization methods. (1) \textbf{GPTQ} \cite{frantar2022gptq} uses a layer-wise quantization technology based on approximate second-order information. (2) \textbf{QuIP} \cite{Quip} adopts a post-training quantization method, which introduces a quantization method of incoherent processing. (3) We also compare with \textbf{OmniQuant} \cite{OmniQuant}, \textbf{AWQ} \cite{AWQ}, \textbf{AQLM} \cite{AQLM}, \textbf{SpQR} \cite{SPQR}, due to space and time limit, we show the results in Appendix \ref{APPD} with only the perplexity results of the OPT model family on the WikiText2 dataset.

\subsubsection{Evaluation Metrics} For perplexity experiments, we use \textbf{\textit{perplexity}} as the metric to evaluate the performance of different models. Lower perplexity indicates that the quantization model is better. For zero-shot experiments, we use \textbf{\textit{accuracy}} to assess the transfer ability of different models. Higher accuracy indicates that the quantization model performs more excellent.

\subsection{Implementation Details} We conduct experiments at three different compression levels: 2-bit, 3-bit and 4-bit. As our method needs to reserve a very small part of important weight parameters ($p\%$ of the weight matrix, and $p = 1$), strictly speaking our method is 2.1-bit, 3.1-bit and 4.1-bit. We select the quantization parameters $S$ and $Z$ using a grid searching strategy between the minimum and the maximum vales of each weight matrix with the goal to minimize $\|\mathbf{W} -\widehat{\mathbf{W}}\|_\text{F}^2$. We set the blocksize $B=128$ consistently. All the experiments are implemented on NVIDIA A800 GPUs and completed within 3 hours. 

\begin{table*}[t]
\centering
\caption{The zero-shot performance of the LLaMA model family on the different datasets (The higher value is better). ``Full''means the full precision method (the original model). For SEPTQ, it is exactly 2.1-bit and 3.1-bit respectively.}
\label{table4}
\tabcolsep=0.4cm
\linespread{2}
\begin{tabular}{l|c|c|c|c|c|c|c|c} 
\toprule
\multirow{2}{*}{Method}     & \multirow{2}{*}{Bits} & \multirow{2}{*}{Size}  &\multicolumn{6}{c}{Accuracy (\%)} \\
\cmidrule(lr){4-9}
            &      &        & PIQA& ARC-e & ARC-c & HellaSwag & Winogrande & Avg.\\
\midrule
   Full        & 16 &\multirow{7}{*}{7B} &77.31          &52.48  &41.38  &72.96      &67.09       &62.24 \\
\cmidrule{1-2} \cmidrule{4-9} 
GPTQ         & \multirow{3}{*}{2} &   &47.88          &25.55  &27.90  &26.09      &48.54       &35.19 \\
QuIP         &  &    &62.13          &36.20  &27.56  &44.76      &51.85       &44.50 \\
\textbf{SEPTQ}&  &      &\textbf{65.83}          &\textbf{40.70}  &\textbf{28.49}  &\textbf{51.16}      &\textbf{58.40}       &\textbf{48.91} \\
\cmidrule{1-2} \cmidrule{4-9} 
   GPTQ     & \multirow{3}{*}{3} &    &73.94          &45.03  &36.43  &65.77      &60.38       &56.31 \\
QuIP         &  &    &75.57          &49.71  &37.63  &68.24      &62.51       &58.73 \\
\textbf{SEPTQ}&  &     &\textbf{76.22}          &\textbf{52.23}  &\textbf{39.33}  &\textbf{69.20}      &\textbf{65.50}       &\textbf{60.50} \\
\midrule
   Full        & 16 &   \multirow{7}{*}{13B}  &79.11          &59.90  &44.53  &76.23      &70.01       &65.96 \\
\cmidrule{1-2} \cmidrule{4-9} 
GPTQ         & \multirow{3}{*}{2} &    &48.97          &25.00  &27.56  &24.90      &52.09       &35.70 \\
QuIP         &  &    &69.42          &44.23  &31.74  &55.47      &56.99       &51.57 \\
\textbf{SEPTQ}& &     &\textbf{74.97}          &\textbf{50.80}  &\textbf{37.12}  &\textbf{66.76}      &\textbf{68.43}       &\textbf{59.62} \\
\cmidrule{1-2} \cmidrule{4-9} 
GPTQ        & \multirow{3}{*}{3} &    &75.52          &52.02  &39.16  &70.38      &65.19       &60.45 \\
QuIP        &  &    &77.20          &\textbf{57.79}  &41.55  &73.23      &67.96       &63.55 \\
\textbf{SEPTQ}& &     &\textbf{78.62}          &57.03  &\textbf{42.06}  &\textbf{74.61}      &\textbf{68.75}       &\textbf{64.21} \\
\midrule
   Full      & 16  &  \multirow{7}{*}{30B}    &80.09          &58.92  &45.48  &79.21     &  72.93    &   67.33   \\
\cmidrule{1-2} \cmidrule{4-9} 
GPTQ    & \multirow{3}{*}{2} &   &50.27          &26.52  &27.73  &25.53      &45.46       &35.10  \\
QuIP         &  &   &71.55          &46.09  &34.22  &61.90      &60.14       &54.78 \\
\textbf{SEPTQ}        &  &    &\textbf{76.99}          &\textbf{54.00}  &\textbf{41.38}  &\textbf{72.09}      &\textbf{70.88}       &\textbf{63.07}  \\
\cmidrule{1-2} \cmidrule{4-9} 
GPTQ         & \multirow{3}{*}{3} &    &77.64          &52.15  &40.44  &74.37      &71.03       &63.13 \\
QuIP       &  &    &77.97          &54.59  &42.91  &75.54      &70.88       &64.38 \\
\textbf{SEPTQ}        &  &     &\textbf{80.30}        &\textbf{57.53}  &\textbf{44.80}  &\textbf{78.21}      &\textbf{73.01}       &\textbf{66.77} \\
\bottomrule
\end{tabular}
\end{table*}

\subsection{Result Analysis of Perplexity Experiments}
Tables \ref{table1}, \ref{table2}, \ref{table3} show the performance of the OPT model family on the C4, WikiText2 and PTB datasets, and the best results are highlighted in bold. According to these tables, it can be seen that in various model sizes, our proposed SEPTQ method obviously outperforms other strong baselines in most cases, especially in extremely low-bit quantization (2-bit). Specifically, when compressing the OPT-66B model to 2 bits on the C4, WikiText2 and PTB datasets, our proposed SEPTQ can achieve 11.96, 9.93 and 14.93 in perplexity respectively, while the perplexity scores of GPTQ and QuIP are only 253.49, 619.32, 630.42 and 17.64, 13.86, 17.50 respectively. It is worth mentioning that the performance of our model at 2-bit level is close to the full precision method. In summary, all these results demonstrate that SEPTQ can compress models across various parameter ranges to very low quantization levels while maintaining robust model performance. 

Strictly speaking, in the above comparison SEPTQ requires to reserve an extra $1\%$ important weight parameters, which seems a little unfair and unreasonable. To further verify the effectiveness of SEPTQ, we set the ratio of important weight parameters to $0.1\%$, i.e., $p=0.1\%$, then our model at 2-bit level becomes a 2.01-bit model. Figure \ref{2bit} shows the perplexity results of the OPT-13B, LLaMA-13B and LLaMA2-13B models on C4 and WikiText2 datasets. For GPTQ and QuIP, they are 2-bit models. And for SEPTQ, it is a 2.01-bit model. From this figure, we can observe that the proposed SEPTQ performs better than GPTQ and QuIP consistently. Specifically, for the results of the LLaMA2-13B model on C4 and WikiText2 datasets, the perplexity scores of SEPTQ are better than those of GPTQ 309.63 and 2090.61 respectively, and better than those of QuIP 7.13 and 5.93 respectively. These results further validate that SEPTQ has the ability to compress the model with satisfactory performance only by reserving a small amount of additional important weight parameters.

\begin{table*}[!t]
\centering
\caption{Running time (in seconds) of different post-training quantization methods on OPT models with WikiText2 at 2 bits.}
\label{table5}
\tabcolsep=0.4cm
\begin{tabular}{l|c|c|c|c|c|c|c|c} 
\toprule
Method  &   125M&   350M&1.3B         & 2.7B   & 6.7B  & 13B    & 30B    & 66B   \\
\midrule
GPTQ &34&91&216 &384	&727	&1311 &2851	&5926  \\
QuIP & 54&127&201&475	&993	&2429	&7788	&16915    \\
SpQR &92&195&376&735	&1440	&2590	&5111	&10354    \\
SEPTQ &43&116&263  &458	&883	&1606	&3126	&6598    \\
\bottomrule
\end{tabular}
\end{table*}

\subsection{Result Analysis of Zero-Shot Experiments}
Table \ref{table4} provides the accuracy results of the LLaMA model family from 7 billions to 30 billions on various datasets including PIQA, ARC-easy, ARC-challenge, HellaSwag, and Winogrande. The best results are highlighted in bold. The experimental outcomes unequivocally demonstrate that our SEPTQ method consistently outperforms other strong baselines across various datasets. Specifically, for the LLaMA-7B, LLaMA-13B, and LLaMA-30B models in the 2-bit quantization level, the SEPTQ method increases the average accuracy by 4.41\%, 8.05\%, and 8.29\% respectively compared with the QuIP method, and increases the average accuracy by 13.72\%, 23.92\% and 27.97\% respectively compared with the GPTQ method. For the LLaMA-13B model on the Winogrande dataset, the accuracy of SEPTQ at the 2-bit level (68.43\%) is even better than those of GPTQ (65.19\%) and QuIP (67.96\%) at the 3-bit level, which further demonstrates the superiority of our proposed SEPTQ method. In addition, it also can be seen that by reserving additional 0.1 bit to record the important weight parameters, the performance gains of SEPTQ on these zero-shot tasks are significant, particularly for 2-bit quantization and larger models. All in all, our proposed SEPTQ has the ability to obtain satisfactory accuracy results while compressing models to very low quantization levels across a broad spectrum of parameters in zero-shot tasks.

\subsection{Comparison of Different Quantization Location Determining Strategies}
(1) For the efficiency, Figure \ref{fig:image1} shows the running time of calculating the important scores for each first layer (including six weight matrices) of the OPT model family by using the static strategy and dynamic strategy respectively. From the results, it can be seen that our proposed static strategy runs much faster than the dynamic strategy across different model sizes. This is because that the dynamic strategy requires to updating weight parameters column-by-column, thus causing additional time cost. (2) For the effectiveness, Figure \ref{fig:image2} shows the perplexity results of the OPT model family on C4, WikiText2 and PTB datasets by using the global strategy and local strategy respectively. From the results, it can be observed that our proposed global strategy consistently performs better than the local strategy across different model sizes. The reason is that the distribution of importance weights is irregular and concentrated in certain areas, thus making it difficult for the local strategy to capture appropriate quantitative locations. (3) Based on the above experiments, we can get that using the static global strategy to determine the quantization location is a good and reasonable choice.

\subsection{Comparison of Running Time among Different PTQ Methods}
The speed of the quantization algorithm is extremely important for post-training quantization. We compared the running time of multiple algorithms, including GPTQ\cite{frantar2022gptq}, QuIP\cite{Quip}, and SpQR\cite{SPQR}. We conduct experiments with OPT on WikiText2 at 2 bits and list the runtime (in seconds) of different PTQ methods as Table \ref{table5}.The results show that our method is a little slower than GPTQ, due to the additional determining the quantization location procedure. But our method is faster than other baselines, especially in larger-scale models like OPT 30B and 66B. In summary, our method can obtain satisfactory performance with acceptable time consumption.

\section{Conclusion}
In this paper, we propose a simple yet effective paradigm for post-training quantization paradigm for large language models, named SEPTQ. The proposed SEPTQ consists of two components: determining the quantization location and quantizing the model weights. For determining the quantization location, we employ a static global strategy which guarantees the effectiveness and efficiency of the method simultaneously. For quantizing the model weights, we leverage the mask matrix that represents the important locations to constrain the derivation of update formula for unquantized weights, thereby precisely  compensate the rounding error caused by quantized weights. Extensive experiments on large language models across different bit levels demonstrate the superiority of our method over other strong baselines. In future work, we plan to explore more accurate quantization location strategies and efficient parameter adaptation methods to improve the current model. 

\section{Acknowledgements}
This work was supported by Liaoning Binhai Laboratory Project (No. LBLF-2023-01), National Natural Science Foundation of China (No. 62106035, 62206038), and Chunhui Project Foundation of the Education Department of China (No. HZKY20220419). We also would like to thank Dalian Ascend AI Computing Center and Dalian Ascend AI Ecosystem Innovation Center for providing inclusive computing power and technical support.

\bibliographystyle{ACM-Reference-Format}
\balance
\bibliography{ref}

\newpage
\appendix

\section{The Detailed Proof Procedure of Eq. (\ref{Eq4})}
\label{APPA}
Here we give the proof procedure from Eq. (\ref{Eq3}) to Eq. (\ref{Eq4}). Given a weight matrix $\mathbf{W}$, the procedure involves quantizing the element at the $i$-th row and $j$-th column, and updating the unquantized weights in the $i$-th row to compensate for the loss. As each row of $\mathbf{W}$ is independent, so altering the values in the $i$-th row does not affect the outputs of the other rows. 

Let the updated $i$-th row $\mathbf{W}_{i,:}$ be $\widehat{\mathbf{W}}_{i,:}$, then Eq. (\ref{Eq3}) can be written as:
\begin{equation}
\begin{split}
\label{Eq7}
s_{i,j}=\text{min}_{\widehat{\mathbf{W}}(i,j)}\|\mathbf{W}_{i,:}\mathbf{X}-\widehat{\mathbf{W}}_{i,:}\mathbf{X}\|_\text{F}^2 \\ \quad \text{s.t.} \quad \widehat{\mathbf{W}}(i,j)=\text{quant}(\mathbf{W}(i,j)),
\end{split}
\end{equation}
where $\|\cdot\|_\text{F}$ is the Frobenius norm, $\widehat{\mathbf{W}}(i,j)$ denotes the matrix $\mathbf{W}$ with only $\mathbf{W}_{i,j}$ is quantized and unquantized elements in the $i$-th row are updated. 

For ease of representation, we use $f(\widehat{\mathbf{W}}_{i,:})=\|\mathbf{W}_{i,:}\mathbf{X}-\widehat{\mathbf{W}}_{i,:}\mathbf{X}\|_\text{F}^2$ to denote the function of $\widehat{\mathbf{W}}_{i,:}$. By leveraging Taylor's expansion on $f(\widehat{\mathbf{W}}_{i,:})$ at $\widehat{\mathbf{W}}_{i,:} = \mathbf{W}_{i,:}$, we can have:
\begin{equation}
\label{Eq8}
\begin{split}
f(\widehat{\mathbf{W}}_{i,:}) &= f(\mathbf{W}_{i,:}) + (\widehat{\mathbf{W}}_{i,:} - \mathbf{W}_{i,:})^T\Delta E(\mathbf{W}_{i,:}) \\
&\quad + \frac{1}{2}(\widehat{\mathbf{W}}_{i,:} - \mathbf{W}_{i,:})^T \mathbf{H} (\widehat{\mathbf{W}}_{i,:} - \mathbf{W}_{i,:}) + o,
\end{split}
\end{equation}
where $\Delta E(\mathbf{W}_{i,:})$ is the first derivative of $f$ with respect to $\widehat{\mathbf{W}}_{i,:}$ at $\widehat{\mathbf{W}}_{i,:} = \mathbf{W}_{i,:}$. The matrix $\mathbf{H}$ is the Hessian matrix which represents the second derivative of $f$ with respect to $\widehat{\mathbf{W}}_{i,:}$ at $\widehat{\mathbf{W}}_{i,:} = \mathbf{W}_{i,:}$, and $o$ is an infinitesimal of higher order.

As $\Delta E(\mathbf{W}_{i,:}) = \textbf{0}$, $f(\mathbf{W}_{i}) =0$ and $\mathbf{H} = 2\mathbf{XX}^T$, then Eq. (\ref{Eq8}) can be written as:
\begin{equation}
\label{eq9}
f(\widehat{\mathbf{W}}_{i,:}) =  \frac{1}{2}(\widehat{\mathbf{W}}_{i,:} - \mathbf{W}_{i,:}) \mathbf{H} (\widehat{\mathbf{W}}_{i,:} - \mathbf{W}_{i,:})^T.
\end{equation}

As $f(\widehat{\mathbf{W}}_{i,:})=\|\mathbf{W}_{i,:}\mathbf{X}-\widehat{\mathbf{W}}_{i,:}\mathbf{X}\|_\text{F}^2$, then substituting Eq. (\ref{eq9} to Eq. (\ref{Eq7}), we can have:
\begin{equation}
\label{Eq10}
\begin{split}
s_{i,j} &= \text{min}_{\widehat{\mathbf{W}}(i,j)} \frac{1}{2}(\widehat{\mathbf{W}}_{i,:} - \mathbf{W}_{i,:}) \mathbf{H} (\widehat{\mathbf{W}}_{i,:} - \mathbf{W}_{i,:})^T, \\
&\quad \text{s.t.} \quad \widehat{\mathbf{W}}(i,j) = \text{quant}(\mathbf{W}(i,j)),
\end{split}
\end{equation}

By using the Lagrange multiplier method, we can have:
\begin{equation}
\label{Eq10}
\begin{split}
s_{i,j} &= \text{min}_{\widehat{\mathbf{W}}_{i,:}} \left[ \frac{1}{2}(\widehat{\mathbf{W}}_{i,:} - \mathbf{W}_{i,:}) \mathbf{H} (\widehat{\mathbf{W}}_{i,:} - \mathbf{W}_{i,:})^T \right. \\
&\quad \left. + \lambda \left(\widehat{\mathbf{W}}(i,j) - \text{quant}(\mathbf{W}(i,j))\right) \right].
\end{split}
\end{equation}

By solving the derivatives of $\widehat{\mathbf{W}}_{i,:}$ and $\lambda$, and make them equal 0, we can get:
\begin{equation}
s_{i,j} = \frac{\left(\mathbf{W}_{i,j} - \text{quant}
(\mathbf{W}_{i,j})\right)^2}{2[\mathbf{XX}^T]^{-1}_{j,j}}.
\end{equation}

\section{The Detailed Proof Procedure of Eq. (\ref{Eq6})}
\label{APPB}
Here we give the proof procedure from Eq. (\ref{OBC}) (with the constraint Eq. (\ref{Eq5})) to Eq. (\ref{Eq6}). The optimization problem posed by the objective function $\text{argmin}_{\mathbf{\widehat{W}}}\|\mathbf{W}\mathbf{X} - \widehat{\mathbf{W}}\mathbf{X}\|_\text{F}^2$ is subject to the constraint $\widehat{\mathbf{W}} = \text{quant}(\mathbf{W}) + \mathbf{M} \odot (\mathbf{W} - \text{quant}(\mathbf{W}))$. For a model with a particularly large weight $\mathbf{W}$ dimension, the time complexity of globally solving the optimal $\widehat{\mathbf{W}}$ is very high. Therefore, we attempt to decompose the problem. First, we rewrite $\|\mathbf{W}\mathbf{X} - \widehat{\mathbf{W}}\mathbf{X}\|_\text{F}^2$ as:
\begin{equation}
\|\mathbf{W}\mathbf{X} - \widehat{\mathbf{W}}\mathbf{X}\|_\text{F}^2 = \sum_{i=1}^{d_{\text{row}}} \|\mathbf{W}_{i,:}\mathbf{X} - \widehat{\mathbf{W}}_{i,:}\mathbf{X}\|_\text{F}^2,
\end{equation}
which allows us to address each row's weight optimization independently. Our focus then shifts to minimize the output difference for each row's weight. Specifically, for $\mathbf{W}_{i,:}$, our goal becomes:
\begin{equation}
\label{eq14}
\text{argmin}_{\widehat{\mathbf{W}}_{i,:}} \|\mathbf{W}_{i,:} \mathbf{X} - \widehat{\mathbf{W}}_{i,:} \mathbf{X}\|_\text{F}^2. 
\end{equation}

According to the proof procedure in Line 1052 - Line 1078, if we introduce the increment $\boldsymbol{\delta} = (\widehat{\mathbf{W}}_{i,:} - \mathbf{W}_{i,:})^T$, where $T$ denotes the transposition, then Eq. (\ref{eq14}) can be written as:
\begin{equation}
    \label{eq15}
  \text{argmin}_{\widehat{\mathbf{W}}_{i,:}} \frac{1}{2} \boldsymbol{\delta}^T  \mathbf{H} \boldsymbol{\delta},  
\end{equation}
where $\mathbf{H} = \mathbf{X} \mathbf{X}^T$ denotes the Hessian matrix. Assuming that we are dealing with the $j$-th value of the $i$-th row, based on Eq. (\ref{Eq5}), we know that $\widehat{\mathbf{W}}_{i,j} = \text{quant}(\mathbf{W}_{i,j}) + \mathbf{M}_{i,j} (\mathbf{W}_{i,j} - \text{quant}(\mathbf{W}_{i,j})),$ where $\mathbf{M}$ is mask matrix of $\mathbf{W}$. Then we can have $\boldsymbol{\delta}_j=\widehat{\mathbf{W}}_{i,j}-\mathbf{W}_{i,j}=\mathbf{e}_j^T  \boldsymbol{\delta}$, where $\mathbf{e}_j$ represents the unit vector with 1 in the $j$-th position. This is to say, we can transfer the constraint of Eq. (\ref{Eq5}) to the constraint:
\begin{equation}
\mathbf{e}_j^T \boldsymbol{\delta}+(\mathbf{W}_{i,j}-\widehat{\mathbf{W}}_{i,j})=0.
\end{equation}

According to Eq. (\ref{eq15}) and the constraint $\mathbf{e}_j^T  \boldsymbol{\delta}+(\mathbf{W}_{i,j}-\widehat{\mathbf{W}}_{i,j})=0$, by using the Lagrange multiplier method, we can have:
\begin{equation}
L(\boldsymbol{\delta}, \lambda) = \frac{1}{2} (\boldsymbol{\delta}^T \mathbf{H}  \boldsymbol{\delta}) + \lambda (\mathbf{e}_j^T  \boldsymbol{\delta} + (\mathbf{W}_{i,j} - \widehat{\mathbf{W}}_{i,j})).
\end{equation}

By solving the derivatives of $\boldsymbol{\delta}$ and $\lambda$, we can get:
\begin{equation}
   \boldsymbol{\delta} = -\frac{\mathbf{W}_{i,j} - \widehat{\mathbf{W}}_{i,j}}{[\mathbf{H}^{-1}]_{jj}}  [\mathbf{H}^{-1}]_{:,j}. 
\end{equation}

\section{Additional Results}
\label{APPD}
We compare our proposed SEPTQ method with several other quantization methods, specifically OmniQuant \cite{OmniQuant}, AWQ \cite{AWQ}, AQLM\cite{AQLM} and SpQR \cite{SPQR}. To ensure clarity and due to space and time limitations, we present the detailed perplexity results of the OPT model family on the WikiText2 dataset. Table \ref{omn1i} summarizes these results, with the best-performing figures highlighted in bold for easy reference. 

We also have attempted to explore whether it is a trick by observing the output projection matrices of some consecutive linear layers (from the 4th to the 7th layers) as Figure \ref{fig:4biterror}. We find that no matter which layer is selected, the similar trend can be observed. 

\begin{table}[b]
\centering
\caption{Quantization bits of SpQR and AQLM.}
\label{SPQRbit}
\tabcolsep=0.20cm
\begin{tabular}{c|c|c|c|c|c|c|c} 
\toprule
 MethodBit     & 125M     & 1.3B  & 2.7B   & 6.7B  & 13B    & 30B    & 66B   \\
\midrule
SpQR(2)    & 2.6   &2.3 &2.3 & 2.1 & 2.2  &  2.1 &2.1  \\
\midrule
SpQR(3)     & 3.7   & 3.5 & 3.5  & 3.5 & 3.3  & 3.5  & 3.9  \\
\midrule
SpQR(4)    & 4.3   & 4.3 & 4.1  & 4.2 & 4.2 &  4.3  &  4.2 \\
\midrule
AQLM(2)    & 2.0   & 2.0 & 1.9  & 2.0 & 2.2 &  1.9  &  1.9 \\
\bottomrule
\end{tabular}
\end{table}

\begin{table*}[t]
\centering
\caption{The perplexity results of the OPT model family on the WikiText2 dataset (The lower value is better). ``Full''means the full precision method (the original model). For SEPTQ, it is exactly 2.1-bit, 3.1-bit and 4.1-bit respectively. Since the number of quantization bits of SpQR and AQLM is flexible, we provide the specific number of quantization bits of SpQR and AQLM for each model in Table \ref{SPQRbit}.}
\label{omn1i}
\tabcolsep=0.5cm
\begin{tabular}{l|c|c|c|c|c|c|c|c} 
\toprule
Method  & Bit     & 125M     & 1.3B  & 2.7B   & 6.7B  & 13B    & 30B    & 66B   \\
\midrule
Full      & 16         & 27.65    & 14.63 & 12.47  & 10.86 & 10.13  & 9.56   & 9.34  \\
\midrule
OmniQuant  & \multirow{5}{*}{2}   & 75.43   & 23.95 & 18.13  & 14.43 & 12.94  & 11.39  & 30.84 \\
AWQ   &      & 251.84   & 47.97 & 28.50  & 16.20 & 14.32  & 12.31  & 14.54 \\
SpQR   &      & 116.94   &22.62 & 16.57  & 14.31 & 11.95  &  \textbf{10.56}  & 9.99 \\
AQLM   &      & \textbf{64.88}   &24.62 & 17.49  &  \textbf{12.91} & \textbf{11.59}  &  10.78  & 10.37 \\
\textbf{SEPTQ}&   & 68.62  & \textbf{20.11} & \textbf{16.20}  & 12.97 & 11.82 & 10.75  & \textbf{9.93} \\
\midrule
OmniQuant   &  \multirow{4}{*}{3}   & 35.66   & 16.68 & 13.80  & 11.65 & 10.87  & 10.00  & 9.83  \\
AWQ   &       & 36.74   & 16.32 & 13.58  & 11.41 & 10.68  & 9.85  & 9.60  \\
SpQR   &       & 34.37   & 15.73 & 12.75  & 11.17 & 10.36  & 9.65 &9.32 \\
\textbf{SEPTQ}&   & \textbf{33.77}   & \textbf{15.46} & \textbf{12.74}  & \textbf{11.14} & \textbf{10.33}  & \textbf{9.61}   & \textbf{9.27}  \\
\midrule
OmniQuant     &   \multirow{4}{*}{4}      & 29.45   & 15.04 & 12.76  & 11.03 & 10.30  & 9.65   & 9.65  \\
AWQ     &       & 32.28   & 15.49 & 12.93 & 11.30 & 10.39  & 9.77   & 9.61  \\
SpQR     &       & 29.37   & 15.03 & 12.57 & \textbf{10.91} & \textbf{10.22}  & \textbf{9.50}  & 9.37  \\
\textbf{SEPTQ} &    & \textbf{29.00}   & \textbf{14.88} & \textbf{12.46}  & 10.98 & \textbf{10.22}  & 9.54   & \textbf{9.27}\\

\bottomrule
\end{tabular}
\end{table*}
\begin{figure*}[t]
    \centering
    \begin{subfigure}{0.88\linewidth}
        \centering
        \includegraphics[width=0.98\linewidth]{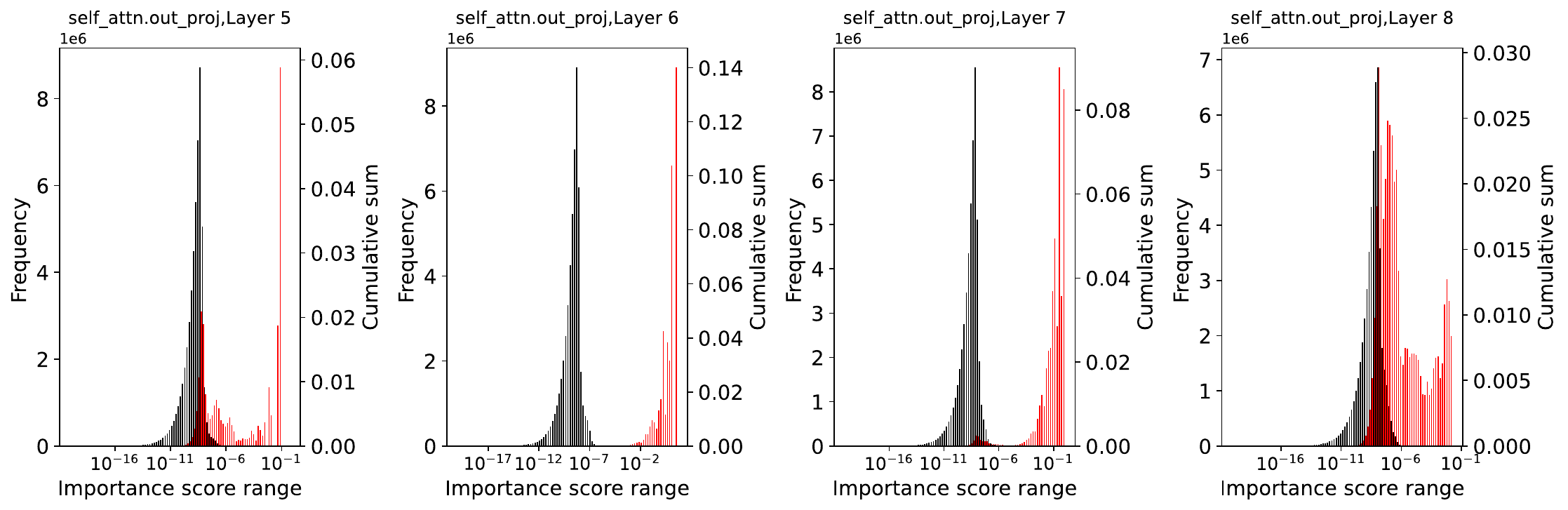}
        \caption{4-bit quantization}
        \label{fig:4biterror}
    \end{subfigure}
    \begin{subfigure}{0.88\linewidth}
        \centering
        \includegraphics[width=0.98\linewidth]{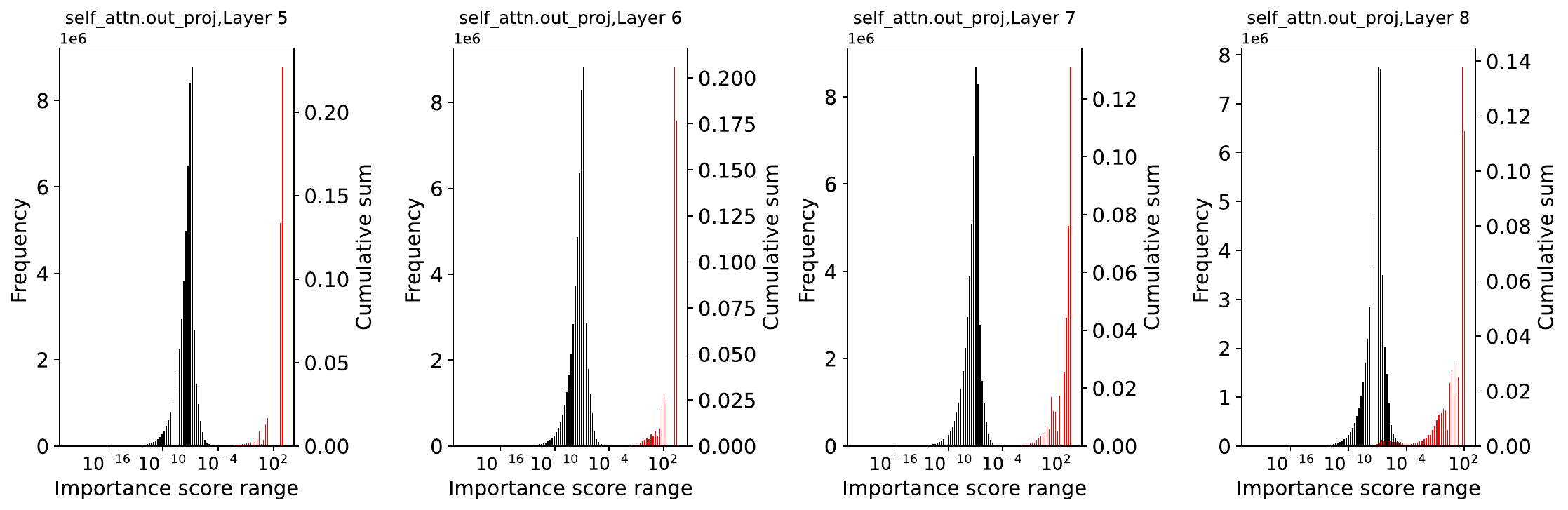}
        \caption{2-bit quantization}
        \label{fig:2biterror}
    \end{subfigure}
    \caption{Output projection matrices of consecutive linear layers (4th to 7th Layers) showing consistent trends across layers}
    \label{fig:4biterror}
\end{figure*}

\end{document}